\documentclass[conference]{IEEEtran}

\usepackage{times}
\usepackage[numbers]{natbib}
\usepackage[bookmarks=true]{hyperref}
\usepackage{amsmath}
\usepackage{relsize}
\usepackage{multicol}
\usepackage{algorithm}
\usepackage{algpseudocode}
\usepackage{graphicx}
\usepackage{subcaption}
\begin{document}

% paper title
\title{Structured Neural Network Dynamics \\ for Model-based Control}

\author{\authorblockN{Alexander Broad\authorrefmark{1}\authorrefmark{2}\authorrefmark{4},
Ian Abraham\authorrefmark{1}\authorrefmark{3},
Todd Murphey\authorrefmark{3}, and
Brenna Argall\authorrefmark{2}\authorrefmark{3}\authorrefmark{4}}
\authorblockA{\authorrefmark{1}These authors contributed equally}
\authorblockA{\authorrefmark{2}Department of Electrical Engineering and Computer Science}
\authorblockA{\authorrefmark{3}Department of Mechanical Engineering\\
Northwestern University, Evanston, IL 60208 \\ 
\authorblockA{\authorrefmark{4}Shirley Ryan AbilityLab, Chicago, IL 60611}
Email: alex.broad@u.northwestern.edu, i-abr@u.northwestern.edu}}

\maketitle

\begin{abstract}
We present a structured neural network architecture that is inspired by linear time-varying dynamical systems. The network is designed to mimic the properties of linear dynamical systems which makes analysis and control simple.  The architecture facilitates the integration of learned system models with gradient-based model predictive control algorithms, and removes the requirement of computing potentially costly derivatives online.  We demonstrate the efficacy of this modeling technique in computing autonomous control policies through evaluation in a variety of standard continuous control domains.
\end{abstract}

\IEEEpeerreviewmaketitle

\section{Introduction and Background}
\label{sec-intro}

The question of how to best generate autonomous control policies for mechanical systems is an important problem in robotics.  Research in this field can be traced back to early work on optimal control by Pontryagin~\cite{pontryagin1987mathematical} and Bellman~\cite{bellman1957dynamic}.  Since this time, significant progress has been made in both the theory and application of autonomous control techniques~\cite{sontag2013mathematical, sutton2017reinforcement}. However, challenges remain in developing strategies that are valid without \textit{a priori} knowledge of the system dynamics.  

One possible solution is to use model-free policy generation techniques~\cite{deisenroth2013survey}.  These methods require no explicit model of the system dynamics and have been shown to be effective in numerous domains~\cite{kober2009policy, levine2014learning}.  However, model-free policy generation techniques often require massive amounts of data and are therefore difficult to evaluate on real-world robotic systems~\cite{deisenroth2013survey}.  An alternative option is to learn an explicit model of the system dynamics which can be incorporated into an optimal control algorithm.  Model-based control methods are more data-efficient and often easier to apply in real-world scenarios~\cite{atkeson1997comparison, broad2017learning}.  However, many optimal control algorithms require some notion of derivatives to compute a control policy~\cite{ansari2016sequential, li2004iterative, mayne1973differential, tassa2014control}.

\begin{figure}[t]
	\centering
	\includegraphics[width=\hsize]{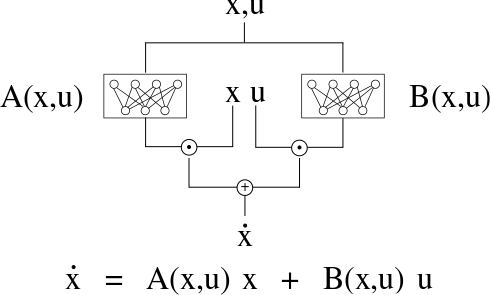}
	\caption{Structured neural network architecture for model-based predictive control.  The final layer of the $A$-subnet must be the same dimension as the state space, and the final layer of the $B$-subnet must be the same dimension as the control space.  The network then computes a function of the form $\dot{x} = A\cdot x + B\cdot u$.  This structure makes it easy to recover time-varying derivatives (i.e., $A, B$) for use in model predictive control algorithms.}
	\label{fig-model}
\end{figure}

Computing the required derivatives can often prove challenging with complex modeling techniques like deep neural networks~\cite{baydin2018automatic}.  Additionally, these black-box methods make it difficult to analyze the underlying dynamics of the system.  There are, of course, alternative modeling techniques~\cite{abraham2017model, atkeson1997comparison, deisenroth2011pilco, gu2016continuous, khansari2011learning, levine2014learning}; however, there remains a desire to incorporate modern, deep neural networks into the optimization loop due to their ability to model challenging dynamic features (e.g., contacts) and scale to high-dimensional tasks~\cite{heess2015learning, nagabandi2017neural, williams2017information}.  In this work, we provide a method that combines the expressive power of neural network models with gradient-based optimal control algorithms.  Our solution is based on a neural network architecture that enforces a linear structure in the state and control space, making it easier to analyze and incorporate into model-based control.   

\section{Structured Neural Networks for \\ Model Predictive Control}
\label{sec-nn-model}

In this section, we define our structured neural network architecture and then detail how the learned models can be integrated into model-based control algorithms. 

\subsection{Structured Neural Network Architecture}
\label{sec-sub-structured-net}

Our neural network architecture is composed of two parallel subnetworks (see Figure~\ref{fig-model}).  The architecture of the first subnetwork ($A$-subnet) can be defined by any number of layers and parameters, and is only constrained such that the final layer must have $N$ parameters, where $N$ is the dimension of the system's state space.  Similarly, the second subnetwork ($B$-subnet) is only constrained such that the final layer must have $M$ parameters, where $M$ is the dimension of the system's control space.  The network then combines (1) the dot product of the output from the $A$-subnet and the state $x$, with (2) the dot product of the output from the $B$-subnet and the control $u$, through an element-wise add operation.  This architecture describes a \textit{single, global model} of the form $\dot{x} = A(x,u) \cdot x + B(x,u) \cdot u$, which is trained with standard gradient-based techniques and can be evaluated and linearized anywhere in the state space.  Here, the $A$-subnet represents the linearization of the dynamics model with respect to the state variables (i.e., $\frac{\partial f}{\partial x}$), and the $B$-subnet represents the linearization of the dynamics model with respect to the control variables (i.e., $\frac{\partial f}{\partial u}$).  

\subsection{Integration with Model-based Control}

Given a learned dynamics model, one can compute autonomous control policies through data-driven methods~\cite{heess2015learning} or through integration with optimal control algorithms~\cite{ansari2016sequential, li2004iterative, tassa2014control}.  On the optimal control side, researchers have mostly explored sampling-based optimization methods.  For example, researchers have proposed computing control trajectories with a random shooting method~\cite{nagabandi2017neural} and with model predictive path integral~\cite{williams2017information} control.  The reason that sampling-based methods are appealing in this domain is that the solution does not depend on computing potentially costly gradients with respect to the state and control variables.  However, the solution does require generating a large number of samples to cover a sufficient portion of the action space.  The challenge, then, is to balance the number of samples generated at each time-step with the rate of the control loop.  As the dimensionality of the action space grows, this becomes more and more challenging.

In contrast with sampling-based methods, gradient-based optimization techniques provide an efficient method of computing control trajectories.  Additionally, these methods provide sensitivity information in the form of time-varying Jacobians.  However, integrating neural network models with these optimization techniques can prove difficult.  This is because it is unclear \textit{a priori} how to compute the necessary Jacobians ($\frac{\partial f}{\partial x}$, and $\frac{\partial f}{\partial u}$).  By enforcing a linear structure on the neural network architecture (as described in Section~\ref{sec-sub-structured-net}), we can efficiently predict the evolution of the dynamic system as well as the required Jacobians.  Then, to generate an autonomous policy, we solve the following optimal control problem
\begin{equation}
	\begin{aligned}
	& \underset{u}{\text{minimize}}
	& & J = \int_{t=0}^{T-1} l(x(t),u(t)) + l_T(x(T)) \\
	& \text{subject to}
	& & \dot{x}(t) = f_\theta(x(t), u(t)), \\
	& & & u(t) \in U, x(t) \in X, \forall t
	\end{aligned}
	\label{eq-mpc}
\end{equation}
\noindent where $f_\theta(x(t),u(t))$ is the learned, structured system dynamics, $l$ and $l_T$ are the running and terminal cost, and $U$ and $X$ are the set of valid control and state values.  The solution of this problem is the control sequence that minimizes the cost. 

\begin{figure*}[t]
  \centering
  \begin{subfigure}{.3\textwidth}
    \centering
    \includegraphics[width=\linewidth]{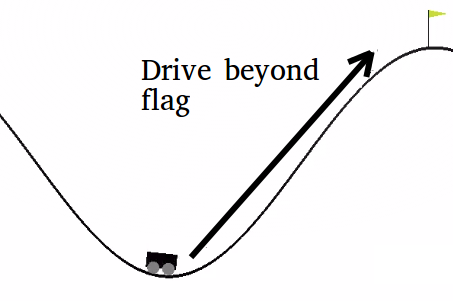}
    \caption{Mountain Car}
    \label{fig-env-1}
  \end{subfigure}
  \begin{subfigure}{.27\textwidth}
    \centering
    \includegraphics[width=\linewidth]{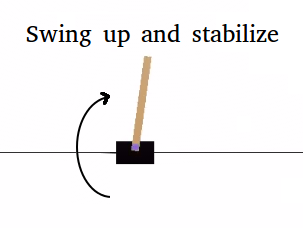}
    \caption{Cart-Pole}
    \label{fig-env-2}
  \end{subfigure}
  \begin{subfigure}{.27\textwidth}
    \centering
    \includegraphics[width=\linewidth]{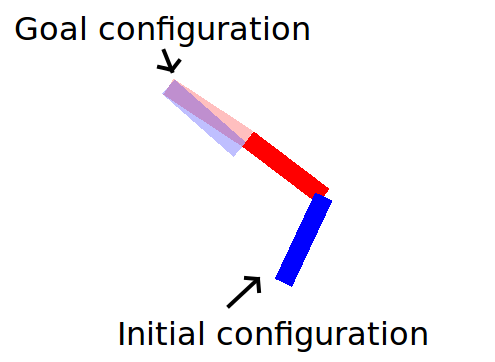}
    \caption{Two-Link Arm}
    \label{fig-env-3}
  \end{subfigure}
  \begin{subfigure}{.3\textwidth}
    \centering
    \includegraphics[width=\linewidth]{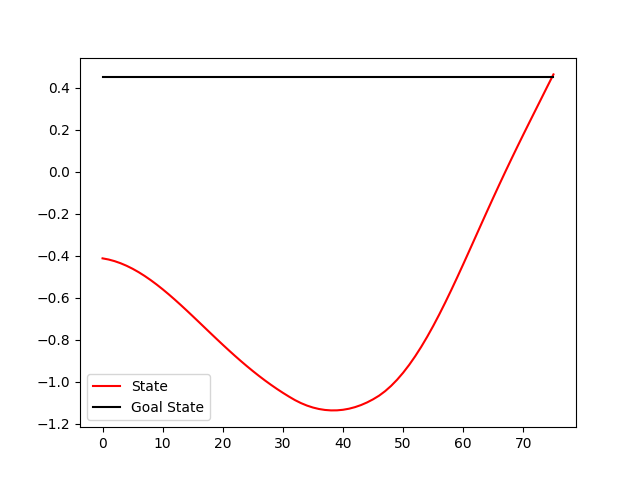}
    \caption{Mountain Car}
    \label{fig-res-1}
  \end{subfigure}
  \begin{subfigure}{.3\textwidth}
    \centering
    \includegraphics[width=\linewidth]{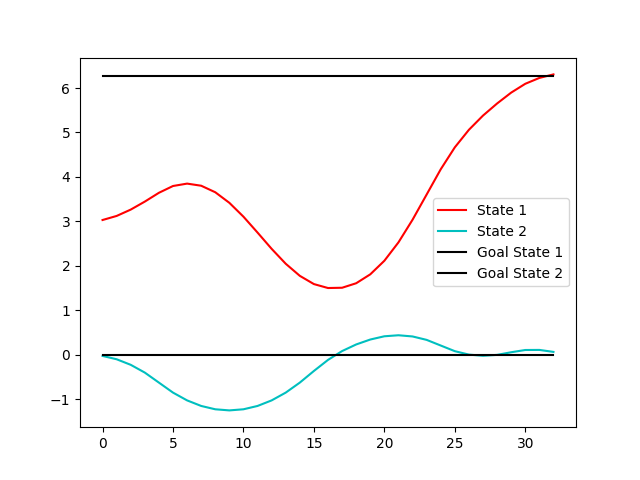}
    \caption{Cart Pole}
    \label{fig-res-2}
  \end{subfigure}
  \begin{subfigure}{.3\textwidth}
    \centering
    \includegraphics[width=\linewidth]{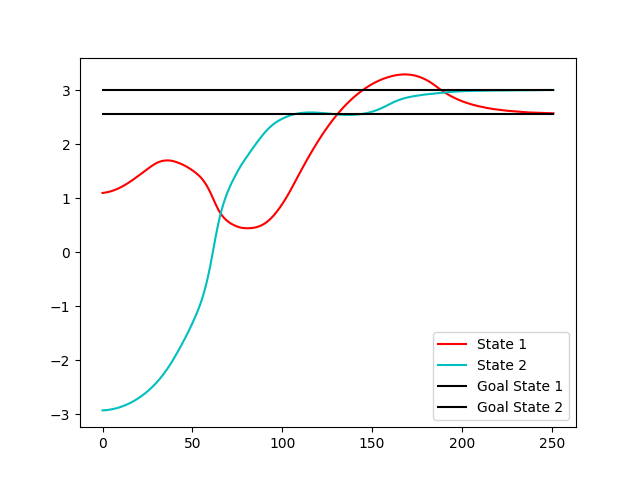}
    \caption{Two Link Arm}
    \label{fig-res-3}
  \end{subfigure}
  \label{fig-envs}
  \caption{Subfigures (a), (b), and (c) are pictorial representations of our experimental environments.  Subfigures (d), (e), and (f) are state diagrams demonstrating the efficacy of our model-based control algorithm.}
\end{figure*}

\section{Experimental Validation}

We validate the efficacy of our described approach through experimentation on three standard control domains.  Our first experimental environment is OpenAI's implementation of the continuous mountain car problem~\cite{brockman2016openai} (Figure~\ref{fig-env-1}).  The mountain car is defined by a two dimensional state space ($x, \dot{x}$) and one dimensional control space ($\ddot{x}$). The second experimental environment is an implementation of the classic cart-pole swing up problem written from scratch (Figure~\ref{fig-env-2}).  The cart-pole is defined by a four dimensional state space ($x, \dot{x}, \theta, \dot{\theta}$) and a one dimensional control space ($\ddot{x}$).  The final experimental environment is a two-link arm written in the Bullet physics engine and described in a related \href{https://katefvision.github.io/}{CMU course} (Figure~\ref{fig-env-3}).  The two-link arm exists in a four dimensional state space ($\theta_1, \dot{\theta_1}, \theta_2, \dot{\theta_2}$) and is controlled with a two dimensional signal ($\ddot{\theta_1}, \ddot{\theta_2}$).  All three environments are defined with continuous-valued state and control spaces.

\subsection{Model Learning Details}
\label{sec-implementation}

In this section, we describe our data collection method and the training procedure.

\subsubsection{Data collection}

We collect data  through observation of trajectories produced by the system using control inputs sampled uniformly at random.  The data is collected in tuples of ($x(t), u(t), \dot{x(t)}$), where $\dot{x}(t)$ is computed as $\frac{x(t) - x(t-1)}{dt}$ and $dt$ is the timestep.  For each environment, we collect 500 trajectories, which are terminated at either 500 steps or when the system violates environment boundary or safety conditions.

\subsubsection{Training the model}

Given a dataset of tuples ($x(t), u(t), \dot{x(t)}$), we train the dynamics model $f_\theta(x(t), u(t))$ by minimizing the following error function
\begin{equation}
	\begin{aligned}
	& & \mathlarger{\varepsilon} = \sum_{(x(t), u(t), \dot{x}(t)) \in D} \frac{1}{2} ||f_\theta(x(t), u(t)) - \dot{x}(t) ||^2
	\end{aligned}
	\label{eq-training}
\end{equation}
We use the Adam optimizer with a learning rate of 0.001.  Half the data is used for training and half for validation.  We find that no data preprocessing is necessary.

\section{Results}

Our evaluation consists of state plots which demonstrate that our defined neural network architecture can be used to solve model-based control problems.  Each example solution depicts the initial state of the system (the start of the state trajectory, which is chosen at random), the time-varying state produced by our model-based control algorithm (red and blue), and the goal state (black).  In Figures~\ref{fig-res-1},~\ref{fig-res-2},~\ref{fig-res-3}, we relay a single solution for each experimental environment, however, we note that our algorithm produced successful control trajectories (with respect to the desired goal state) from a variety of initial conditions.  Additionally, our approach was able to successfully generate control trajectories that reached arbitrary goal states in the two-link arm environment.  

These results suggest that our structured neural network can be used to learn a global model of the system dynamics, while simultaneously enforcing linearization constraints that make it possible to recover time-varying derivatives without additional computation.  In contrast to approximation methods (e.g., numerical differentiation) and symbolic methods (e.g., automatic differentiation), our approach can be thought of as a \textit{prediction method} for computing the required time-varying derivatives.  Related work in this area includes the \textit{transformation network} proposed in~\cite{watter2015embed} which directly predicts the parameters of an A and B matrix in a latent space.  In contrast, our approach does not explicitly learn parameters of a matrix; instead we learn nonlinear mappings (A-subnet, B-subnet) \textit{that we treat as linearizations} of the global model in the structure of our network network.  This allows us to learn a global model of the system dynamics, while simultaneously enforcing linearization constraints.  A related call for the use of structure in neural networks has been explored in model-free policy generation.  In~\cite{srouji2018structured}, researchers describe a network architecture that combines linear and nonlinear policies into a single control model.  In our work, we instead enforce structure that mimics linear time-varying systems, and incorporate these models into optimal control algorithms.  

\subsection{Why We Think This Works}

In this work, we address the bottleneck associated with computing gradients of the system model through the application of a \textit{structured neural network} that explicitly encodes linearization constraints and therefore reduces the computational complexity necessary to recover the required Jacobians.  However, without further study, it is not clear whether or not the learned $A$ and $B$-subnetworks actually approximate the required time-varying derivatives.  Experimental evidence suggests that the vectors represented by these networks are, at a minimum, pointing in the direction of the gradient.  This claim is based on the fact that (1) our model-based control algorithm produces successful policies in a variety of control domains, and (2) when we incorporate the learned system model into an MPC algorithm, we treat the output of the subnetworks as first order derivatives of the system dynamics.

\subsection{Open Questions}

We now pose a number of open questions that we plan to address in future work.  In particular, we are interested in exploring how our structured neural network model compares with alternative methods of computing time-varying derivatives.  One such solution is to use a finite differences method for numerical differentiation.  From a practical standpoint, we note that this method is prone to round-off errors and is computationally expensive in an iterative, receding-horizon framework.  Another solution is to use automatic differentiation~\cite{baydin2018automatic}.  This approach has been shown to work well, however it requires well formed expression graphs and derivatives computed at compile-time to work efficiently enough for online optimization~\cite{giftthaler2017automatic}.  In future work, we plan to compare and contrast these methods in high-dimensional control spaces.  

\section{Conclusion}

In this work, we propose a structured neural network that can be used to solve model-based control problems.  The architecture makes it easy to integrate the learned models with gradient-based optimal control algorithms and simplifies the interpretation of a system model parameterized by a deep neural network.  This idea is inline with other recent calls for simplification of data-driven control strategies such as~\cite{mania2018simple, rajeswaran2017towards}.  %Additionally, we provide preliminary evidence of the efficacy of approach in some simple continuous control domains.  

\bibliographystyle{plainnat}
\bibliography{references}

\end{document}